\documentclass[journal,comsoc,draftnocls,onecolumn]{IEEEtran}

\ifCLASSINFOpdf
\else
\fi

\usepackage{amsmath}
\usepackage{graphicx}
\usepackage{amssymb}
\usepackage{cite}
\usepackage[cmintegrals]{newtxmath}

\usepackage[english]{babel}
\usepackage[utf8x]{inputenc}
\usepackage[T1]{fontenc}

\usepackage{colortbl}
\usepackage[font={small}]{caption}
\usepackage[colorinlistoftodos]{todonotes}
\usepackage[colorlinks=true, allcolors=blue]{hyperref}
\usepackage[keeplastbox]{flushend}
\usepackage{acronym}
\usepackage{dsfont}
\usepackage{color}
\usepackage{multirow}
\usepackage{lastpage,algorithm,algorithmic}
\usepackage{epsfig,amssymb}
\usepackage{bbding}
\usepackage{booktabs}
\usepackage{color, balance}

\begin{document}
%
% paper title
% Titles are generally capitalized except for words such as a, an, and, as,
% at, but, by, for, in, nor, of, on, or, the, to and up, which are usually
% not capitalized unless they are the first or last word of the title.
% Linebreaks \\ can be used within to get better formatting as desired.
% Do not put math or special symbols in the title.
\title{\huge{When Multiple Agents Learn to Schedule: A Distributed Radio Resource Management Framework}}
%
%
% author names and IEEE memberships
% note positions of commas and nonbreaking spaces ( ~ ) LaTeX will not break
% a structure at a ~ so this keeps an author's name from being broken across
% two lines.
% use \thanks{} to gain access to the first footnote area
% a separate \thanks must be used for each paragraph as LaTeX2e's \thanks
% was not built to handle multiple paragraphs
%

%\iffalse
\author{Navid Naderializadeh, Jaroslaw Sydir, Meryem Simsek, Hosein Nikopour, and Shilpa Talwar% <-this % stops a space
\thanks{The authors are with Intel Corporation, Santa Clara, CA 95054, USA. E-mails: $\{$navid.naderializadeh, jerry.sydir, meryem.simsek, hosein.nikopour, shilpa.talwar$\}$@intel.com.% <-this % stops a space
}}
%\fi

% make the title area
\maketitle

% As a general rule, do not put math, special symbols or citations
% in the abstract or keywords.
\begin{abstract}
Interference among concurrent transmissions in a wireless network is a key factor limiting the system performance. One way to alleviate this problem is to manage the radio resources in order to maximize either the average or the worst-case performance. However, joint consideration of both metrics is often neglected as they are competing in nature. In this article, a mechanism for radio resource management using multi-agent deep reinforcement learning (RL) is proposed, which strikes the right trade-off between maximizing the average and the $5^{th}$ percentile user throughput. Each transmitter in the network is equipped with a deep RL agent, receiving partial observations from the network (e.g., channel quality, interference level, etc.) and deciding whether to be active or inactive at each scheduling interval for given radio resources, a process referred to as link scheduling. Based on the actions of all agents, the network emits a reward to the agents, indicating how good their joint decisions were. The proposed framework enables the agents to make decisions in a distributed manner, and the reward is designed in such a way that the agents strive to guarantee a minimum performance, leading to a fair resource allocation among all users across the network. Simulation results demonstrate the superiority of our approach compared to decentralized baselines in terms of average and $5^{th}$ percentile user throughput, while achieving performance close to that of a centralized exhaustive search approach. Moreover, the proposed framework is robust to mismatches between training and testing scenarios. In particular, it is shown that an agent trained on a network with low transmitter density maintains its performance and outperforms the baselines when deployed in a network with a higher transmitter density. Finally, challenges and ramifications of employing the proposed framework in real-world networks are discussed.
\end{abstract}

% Note that keywords are not normally used for peerreview papers.

\IEEEpeerreviewmaketitle

\section{Introduction}
One of the key drivers for improving throughput in future wireless networks, including fifth generation mobile networks (5G), is the densification achieved by adding more base stations. However, the rise of such ultra-dense network paradigms implies that the wireless resources (in time, frequency, etc.) need to support an increasing number of transmissions. Effective radio resource allocation procedures are, therefore, critical to mitigate interference and achieve the desired performance enhancement in this ultra-dense environment.

The radio resource allocation problem is in general non-convex and therefore computationally complex, especially as the network density increases. Researchers have devised a spectrum of centralized and distributed algorithms for radio resource allocation, using various techniques such as geometric programming~\cite{gjendemsjo2008binary}, weighted minimum mean square optimization~\cite{shi2011iteratively}, information-theoretic optimality of treating interference as noise~\cite{naderializadeh2014itlinq}, and fractional programming~\cite{shen2017fplinq}. However, these heuristic algorithms may sometimes fail due to the dynamics of the wireless environment. Such dynamics may better be handled by algorithms that learn from interactions with the environment. In fact, solutions based on machine learning, and in particular reinforcement learning (RL), have emerged as an alternative to better optimize various aspects of resource allocation in wireless systems~\cite{stabellini2008interference, Simsek, nasir2018deep, eisen2019learning, ahmed2019deep}.

A common drawback of most prior works, in this context, is that they intend to optimize a single metric or objective function, a prominent example of which is the sum-throughput of the users across the network. However, resource allocation solutions which optimize the sum-throughput often allocate resources unfairly across the users, as they only focus on the average performance and fail to guarantee a minimum performance. Moreover, many learning-based solutions proposed in the literature do not address the possible mismatch between the training and deployment environments and the robustness of the solution to changes in the environment is not considered.

In this article, the application of deep RL techniques to the problem of distributed interference mitigation in multi-cell wireless networks is discussed, and a mechanism for scheduling transmissions using deep RL is proposed so as to be fair across all users throughout the network. In particular, each transmitter in the network is considered to be a deep RL agent. A centralized training mechanism is utilized to train all agents, which are competing for the same wireless resource, to maximize the network throughput, while at the same time attempting to maintain a minimum user throughput in order to achieve fairness across all users. The agents are trained to make their link scheduling decisions in a distributed fashion, relying only on a local set of observations from the wireless environment (such as the local channel quality and received interference levels), alongside some observations obtained from a subset of neighboring agents. Observations obtained from neighboring agents are sampled less frequently and delayed to match the constraints imposed by real-world implementations.

The proposed deep RL-based link scheduling mechanism is evaluated using a system-level simulator and its performance is compared against two baseline approaches, and a genie-aided centralized exhaustive-search approach. The simulation results demonstrate that the proposed approach outperforms both baselines in terms of average user throughput and the $5^{th}$ percentile user throughput. Moreover, the proposed approach achieves a performance close to that of exhaustive search. 

It is also shown that the introduced framework is robust to changes in the environment. The proposed structure for the deep RL agent allows a trained model on one specific setting to be applicable in a wide range of different scenarios. Specifically, a deep RL agent is trained for a scenario with a low number of transmitters and, upon the completion of training, tested across scenarios with higher numbers of transmitters. It is observed that the agent can maintain a decent level of performance in all cases, hence implying the scalability and robustness of the proposed framework with respect to the density of nodes in the network.

Throughout the article, challenges to use the proposed deep RL framework in a real-world wireless network are highlighted. An end-to-end architecture for applying different stages of this framework is provided. In particular, practical solutions are discussed for training, deployment, and information exchange among different entities in the network.

\section{Background on Radio Resource Allocation and Deep Reinforcement Learning}

In this section, the general problem of wireless resource allocation is outlined, and then a high-level overview of multi-agent deep reinforcement learning is presented.

\subsection{Radio Resource Management in Wireless Networks}

\begin{figure}[t]
\centering
\includegraphics[trim = .6in 2.9in .5in 3.1in, clip,width=0.98\textwidth]{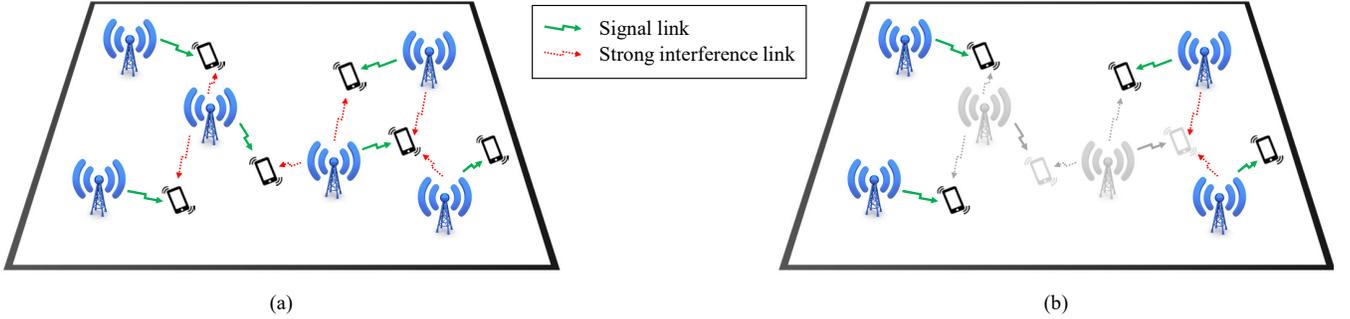}
\caption{(a) A wireless network with multiple AP-UE pairs, and (b) impact of link scheduling on the network, where only a subset of APs are activated simultaneously so as to mitigate the interference among concurrent transmissions.}
\label{fig:network}
\end{figure}

A wireless network is considered, as in Figure~\ref{fig:network}-a, comprising multiple access points (APs) and multiple user equipment nodes (UEs) in the downlink mode, where the APs intend to serve the UEs over multiple scheduling intervals. Before the data transmissions begin, each UE is associated with the AP to which it has the strongest long-term channel gain. Afterwards, the network runs for many scheduling intervals, at each of which, each AP selects one of its associated UEs to which it will transmit data in that interval.

As all transmissions use the same wireless spectrum, APs which simultaneously serve their corresponding UEs interfere with each other. In such a scenario, one way to alleviate interference is to activate only a subset of transmissions at each scheduling interval, so that they cause a reduced level of interference to each other. This process is called link scheduling, as illustrated in Figure~\ref{fig:network}-b. The objectives of link scheduling are two-fold: Maximizing the average rate (equivalently, sum-rate) and maximizing the $5^{th}$ percentile rate across all UEs over all instances of the environment (placements of APs and UEs). These two objectives are in natural tension with each other, as maximizing average user rate can cause users experiencing bad channel conditions to receive no transmissions, while focusing only on the worst-case users may reduce the average rate. Therefore, the goal is to strike the right balance between these two metrics.

In order to enable the APs to make link scheduling decisions, the UEs periodically measure and report back their channel conditions through channel quality indicator (CQI) feedback reports~\cite{3gpp.38.214}. The APs can rely on such reports, as well as other local information from the network, to make their scheduling decisions. Moreover, the APs can communicate their local information with each other in order to provide each other with a better understanding of the environment.

\subsection{Overview of Multi-Agent Deep Reinforcement Learning}

\begin{figure}[t]
\centering
\includegraphics[trim = 1.6in 2.9in 1.9in 3.2in, clip,width=0.98\textwidth]{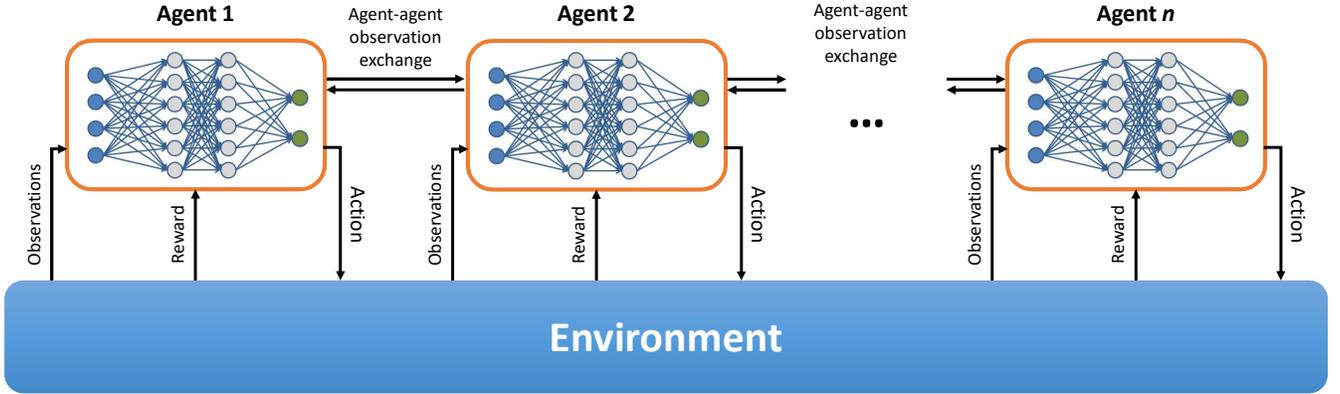}
\caption{Multi-agent deep reinforcement learning diagram, where the agents are allowed to exchange their observations with neighboring agents. The rewards emitted to the agents allow them to train their policies in order to take actions that maximize their long-term rewards over time.}
\label{fig:MARL}
\end{figure}

The multi-agent RL paradigm consists of multiple agents, interacting with each other and with an environment, as illustrated in Figure~\ref{fig:MARL}. Each agent can either observe the full state of the environment or make only a partial observation. At each time step, the agents receive observations from the environment, based on which they take actions from the set of feasible actions allowed for each agent. These actions make the environment transition to a new state and as a result, the environment emits a reward to each agent.  The reward received by each agent may be local to that agent, indicating the impact of the agents’ actions on its local objective, or global to all agents, indicating the impact of the agents’ actions on their collective objective, or a combination of the two. The goal for each agent is to learn to take actions so as to maximize its cumulative future reward.

In this article, each agent is considered to be equipped with a deep Q-network (DQN)~\cite{mnih2013playing}, which is a deep neural network (DNN) through which the agent decides on the action it should take based on its current observations. In particular, the number of input nodes in the DQN is equal to the length of the observation vector of each agent, and the number of output nodes is equal to the number of its possible actions. At each time step, each agent feeds its observation vector to its DQN, and at the output, it derives the value of each action given the current observations. It can then exploit the DQN by taking the action with the maximum value, or it can explore by ignoring the DQN entirely and taking a random action. A trained DQN is referred to as a developed policy, according to which an agent selects the best action to perform at each step.

The DQNs that are trained for and used by each of the agents can be identical or distinct. In this work, we consider a single DQN that is shared among all the agents. This implies that centralized training is performed by collecting the experiences of all agents and using them to train a single DQN, which is then used by all the agents. Even though training is centralized, the execution phase happens in a completely distributed way, with each agent making its own decision at each time step only based on the specific observations it receives from the environment, which can include some delayed observations from neighboring agents. Stated differently, the system learns one policy that is used by all the agents, but the agents use their individual observations to decide on their individual actions according to this common policy. Later in the article, the practical implications of a centralized training mechanism will be demonstrated in a real-world deployment scenario.

\section{Proposed Framework for Applying Deep RL to Radio Resource Allocation}

In this section, some of the challenges of applying deep RL to wireless network resource allocation is discussed. Afterwards, the details of the proposed link scheduling algorithm using multi-agent deep RL are presented, starting with the description of the observations, actions and rewards and then the entire training methodology.

\subsection{Challenges}

Unlike many of the environments to which deep RL has been applied, for example Atari games~\cite{mnih2013playing}, the definition of the state space and reward for the problem of radio resource allocation is not obvious. Agents have access to various measured indicators of channel quality and interference levels, such as the signal-to-interference-plus-noise-ratio (SINR) values reported by their associated UEs. The APs can calculate the long-term average rates achieved by their UEs and estimate their proportional-fair (PF) ratio, as well. They may also remember their past actions and observations. It is feasible for agents to communicate and share their observations with each other, assuming that the delays in those information exchanges are accounted for. 

The formulation of the reward function is a second important challenge. Rewards can be based on a number of centralized measures of network-wide performance, such as sum-rate or the product of the rates. Rewards can be local to each agent, based on the rates achieved by the UEs associated with the agent’s AP, or global, indicating the performance across all APs/UEs in the network. The selection of which observations and reward(s) to utilize are critical choices, which have a significant impact on the learning process and system performance.

Due to the nature of wireless networks, the density and relative placement of APs and UEs may vary greatly from deployment to deployment and even within a single deployment, and the training procedure must ensure that the space of all possible AP/UE locations and channel realizations is covered to enable the learned policy to operate in a broad range of possible scenarios. A trade-off needs to be made between the number of training steps taken against each realization of the environment and the number of environment realizations used in training. Finally, the performance of the algorithm as it evolves during the training procedure must be evaluated in a comparable way in order to determine whether learning is happening successfully, to select the best model, and to determine when learning should be terminated.

\subsection{Structure of Observations, Actions, and Rewards}

In what follows, the observations, actions, and rewards for each of the agents are described in detail.

\begin{itemize}
\item \textbf{Observations:} The observations of each agent consist of a set of local observations and a set of remote observations, which are received from a subset of neighboring agents. The neighboring agents of an agent are defined as the concatenation of the three APs that cause the strongest interference at the UE associated with the agent, and the three APs whose associated UE suffers from the strongest interference caused by this agent. Both local and remote observations are assumed to include the measured SINRs by the corresponding UEs, alongside the weights of those UEs, defined as the inverse of the UEs’ long-term average rates. The long-term average rates of the UEs are calculated using an exponential moving average, allowing them to adapt to recent rates experienced by the UEs in a smooth fashion. The observations are periodically updated with delayed information, reflecting the inherent delays in reporting and processing the CQI feedback, as well as the communication delay between the agents.

\item \textbf{Actions:} The action of each agent consists of a binary decision, indicating whether it should be active/inactive at each scheduling interval.

\item \textbf{Rewards:} A centralized reward is emitted to each of the agents in the form of weighted sum-rate of the UEs across the network at the corresponding scheduling interval, where the weights of the UEs are defined as mentioned above. Moreover, in case all agents decide to be inactive, resulting in a zero weighted sum-rate, the agent whose UE has the highest PF ratio receives a negative reward equal to the negative of that PF ratio, while the rest of the agents receive zero reward. This reward mechanism ensures that the agents are encouraged to not all turn off at the same time. Furthermore, the weighted sum-rate reward encourages a balanced resource allocation, as the UEs in worse conditions have lower long-term average rates and higher weights, which if served, lead to higher rewards.
\end{itemize}

\subsection{Training Methodology}

\begin{figure}[t]
\centering
\includegraphics[trim = .7in 2.3in 1.3in 2.5in, clip,width=0.98\textwidth]{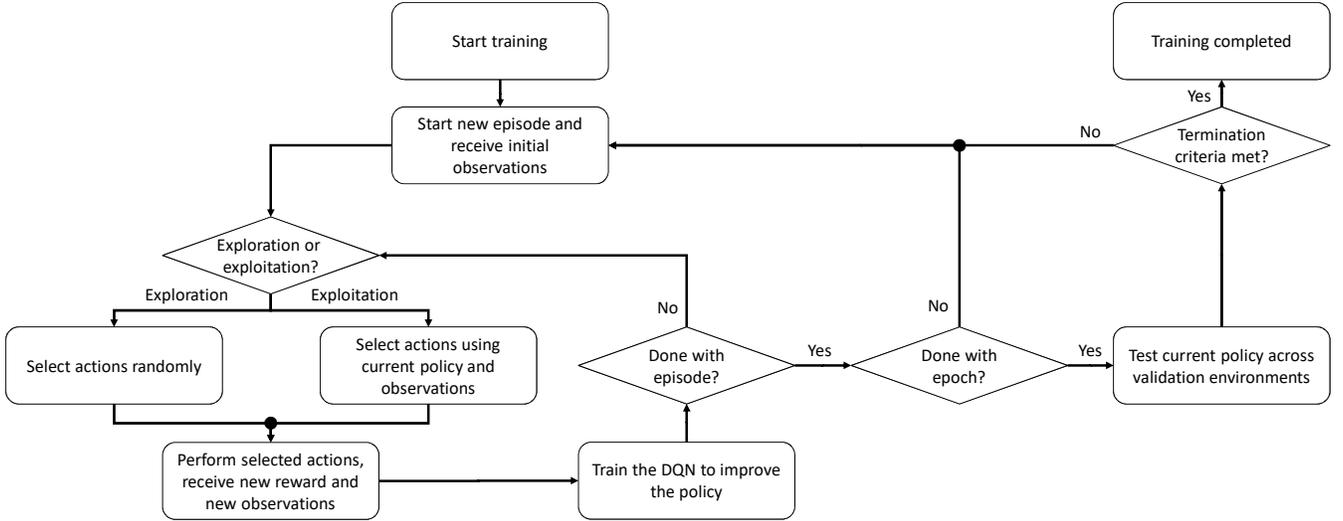}
\caption{The proposed end-to-end architecture for training deep RL agents to optimize radio resource allocation.}
\label{fig:flowchart}
\end{figure}

Our training procedure, as depicted in Figure~\ref{fig:flowchart}, is divided into episodes. Each episode is performed on a realization of the environment in which the locations of the APs and UEs and channel realizations are randomly selected following a set of rules on minimum AP-AP and UE-AP distances. An episode consists of a fixed number of scheduling intervals, where at each interval, the agents decide on which actions the APs should take. The experiences gained by all of the agents are gathered and used as inputs for training. The exploration of the state-action space is driven by two mechanisms. Across episodes, the selection of AP/UE locations and generation of the shadowing and short-term fading realization drives state exploration. On the other hand, within a specific episode, an $\epsilon$-greedy type of approach for selecting actions either by exploiting the policy or randomly drives action exploration.

The training is further structured into epochs, each of which consists of multiple episodes. At the completion of each epoch, training is paused in order to evaluate the current policy against a fixed set of validation environments. The same set of validation environments are used at the end of each epoch and they start from the same initial state every time. This allows the performance of the evolving policy to be tracked epoch-by-epoch using a measure combining average UE rate and $5^{th}$ percentile UE rate that is indicative of the performance across a broad range of possible environments. The performance measures obtained at the end of each epoch are used to select the best model and to determine when training should be terminated because the model performance has converged.

The validation environments are carefully selected to represent a broad range of possible environments, while not being too large, so as to minimize training time. Through experimentation, a set of 1000 randomly selected environments was found to provide a good representation of all possible environments in terms of average rate and $5^{th}$ percentile rate using a simple baseline link scheduling scheme. Afterwards, a set of validation environments is randomly created, and the average rate and $5^{th}$ percentile rate are calculated using the same link scheduling scheme and compared to the values derived from the 1000 environment set. This process is repeated until the values are within a negligible relative error (e.g., 5\%).

\section{Simulation Results}
In this section, the performance of the proposed algorithm is demonstrated through system-level simulation results and compared with multiple baseline algorithms. A network with 4-10 APs is considered, with each AP having a single UE associated with it. At the beginning of each episode, each of the APs is dropped uniformly at random within a 500m×500m area by considering a minimum distance constraint of 35m between them. After the AP locations are set, for each AP, a UE is dropped randomly in a circle around the AP with a radius of 100m, ensuring a minimum AP-UE distance threshold of 10m. Each UE is then associated with the AP to which it has the strongest long-term channel gain, consisting of a dual-slope path-loss model~\cite{zhang2015downlink} and 7dB log-normal shadowing. The sum of sinusoids (SoS) model~\cite{li2002simulation} is used to implement short-term flat Rayleigh fading channel. The transmit power of each AP is assumed to be 10 dBm, the noise power spectral density is taken to be -174 dBm/Hz, and a bandwidth of 10 MHz is considered. Each episode is run for a total of 400 scheduling intervals, the first 200 of which are used to stabilize the observations, while the rest are used to train and evaluate the performance.

As for the agent, a double DQN architecture is used~\cite{van2016deep}, which involves a 2-layer neural network with 128 neurons per layer and a tanh activation function. The agents are initialized with 100 pre-training episodes, in which they only take random actions to explore the environment. The exploration probability then decreases linearly to 1\% within 50 episodes to gradually let the agents exploit the trained policy. The main DQN is updated every 20 scheduling intervals, using a batch of 64 sample time instances from the replay buffer. To account for the correlation between agents’ observations and actions, the samples are drawn using concurrent sampling~\cite{omidshafiei2017deep}. The target DQN, however, is updated 100 times less frequently to stabilize training. The training is run for 2000 epochs, each of which consists of 50 episodes. At the end of each epoch, the developed policy is evaluated across a set of 50 validation environments to assess its learning curve and decide whether or not to save the model.

\begin{figure}[t]
\centering
\includegraphics[width=\textwidth]{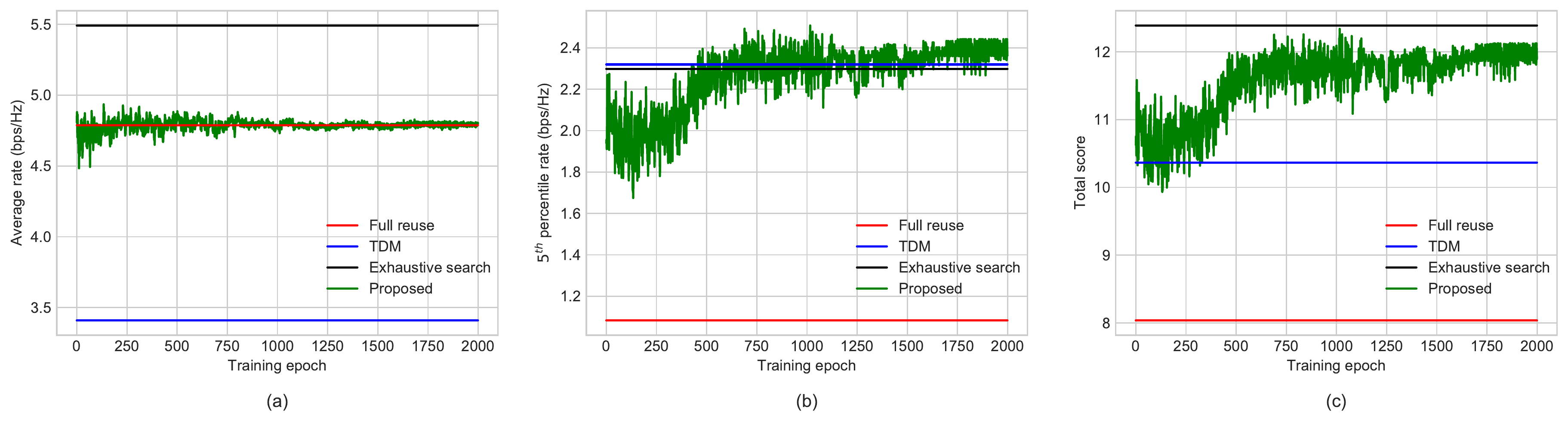}
\caption{Evolution of (a) average rate, (b) $5^{th}$ percentile rate, and (c) total score over time during training, where the total score is a linear combination of average rate and $5^{th}$ percentile rate, emphasizing on the latter metric.}
\label{fig:evolution_over_time}
\end{figure}

As mentioned before, inherent delays are considered for information exchange between the UEs and APs, and also between different APs. In particular, it is assumed that the local observations are updated every 10 scheduling intervals and become available to the agent after a delay of 4 intervals. The remote observations are also updated every 10 intervals. However, they are available only after a delay of 20 intervals to account for the information exchange latency between agents.

The performance of the proposed algorithm is compared against the following baselines:

\begin{itemize}
\item \textbf{Full reuse:} All agents are active at all scheduling intervals.
\item \textbf{TDM:} The agents take turns for transmission in a round robin fashion, with only one of them being active at each interval.
\item \textbf{Exhaustive search:} At each interval, a central entity decides on the set of actions for all APs that maximizes the weighted sum-rate of the UEs in that interval.

\end{itemize}

Figure~\ref{fig:evolution_over_time} shows the performance of the proposed approach during training, evaluated on the 50 validation environments for a network of 4 AP-UE pairs. As the figure illustrates, the model performance improves as the training proceeds both in terms of average rate and the $5^{th}$ percentile rate. In order to assess the trade-off between these two metrics, a total score is defined as a specific linear combination of average rate and $5^{th}$ percentile rate, which weighs the latter three times more than the former. Figure~\ref{fig:evolution_over_time}-c demonstrates that the proposed approach achieves a higher score than both full reuse and TDM, and can almost attain the same score as exhaustive search. The model is saved every time a new maximum score is encountered during training.

\begin{figure}[t]
\centering
\includegraphics[width=0.6\textwidth]{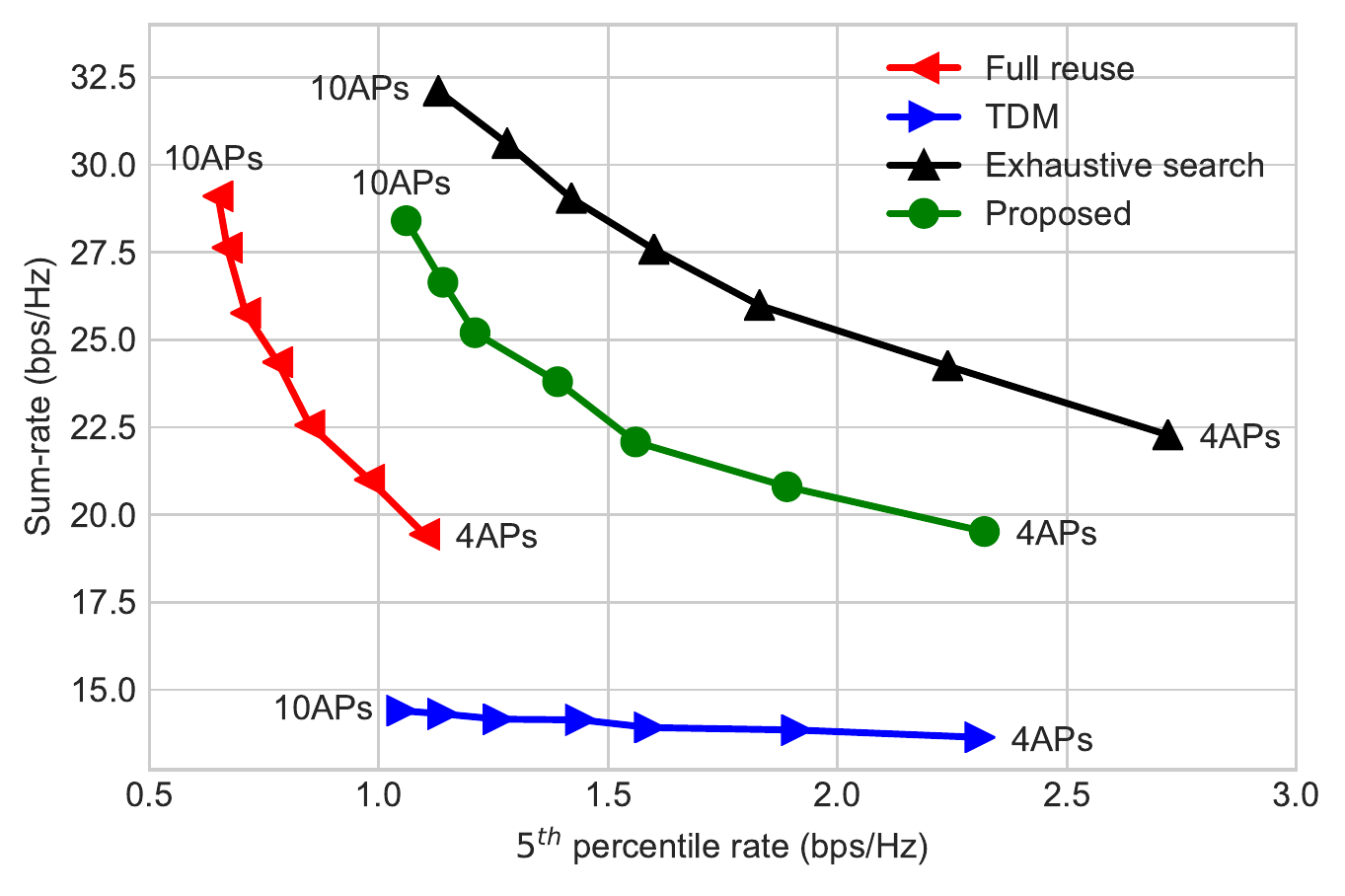}
\caption{Trade-off between sum-rate and $5^{th}$ percentile rate achieved by the proposed approach and the baselines for networks with 4-10 APs. Full reuse suffers from low $5^{th}$ percentile rate, while TDM hurts average rate by blindly dividing resources across all links. The proposed approach, however, strikes the right balance between these two metrics, attaining a similar performance to that of exhaustive search.}
\label{fig:sumrate_vs_coverage}
\end{figure}

In order to conduct a final test after training is complete, the best trained model (saved at the absolute highest maximum score) is tested across 1000 random environment instances with 4-10 APs. Figure~\ref{fig:sumrate_vs_coverage} shows the achieved (sum-rate, $5^{th}$ percentile rate) pairs by the proposed DQN scheduler and the baseline schemes. As the figure demonstrates, while both full reuse and TDM schemes exhibit poor trade-offs between sum-rate and $5^{th}$ percentile rate, the proposed approach strikes a much better balance compared to those two schemes. It also achieves a performance close to that of exhaustive search, while being executed in a distributed fashion as opposed to exhaustive search, which is centralized and computationally prohibitive. Moreover, note that the DQN agent was trained on a network with only 4 APs. Therefore, Figure~\ref{fig:sumrate_vs_coverage} also demonstrates the scalability and robustness of the proposed approach. In particular, the introduced design of the observation space enables the trained policy to be applicable to a broad range of densities in the network. Consequently, in case the network density during deployment is not the same as the network density during training, not only can our trained agents still be deployed, but they also enjoy a decent performance.

\section{Limitations and Implications in Real-World Deployments}
In order for deep RL-based algorithms to be practical in real-world networks, the training procedure must learn a policy (or a set of policies) which produces a reasonable behavior when deployed in a real network. Even if additional training is expected to occur after deployment, the initial policy must be good enough to operate in the network even if the performance is not optimal. The goal of the training procedure is to produce such a policy using a simulation of the wireless network environment and dynamics. Specifically, the training should aim to develop a policy that works for a wide variety of environments, with different AP/UE densities, different configurations in terms of AP/UE locations, and different channel realizations.

Considering a practical deployment of APs and UEs in a wireless network, where each AP is equipped with an agent, the diagram in Figure~\ref{fig:real} shows the end-to-end architecture of training and inference that may take place in a system with, for example, two AP-UE pairs. The links in the diagram can be decomposed to two subsets:

\begin{figure}[t]
\centering
\includegraphics[trim = 2.15in 2.2in 3.05in 2.4in, clip,width=0.75\textwidth]{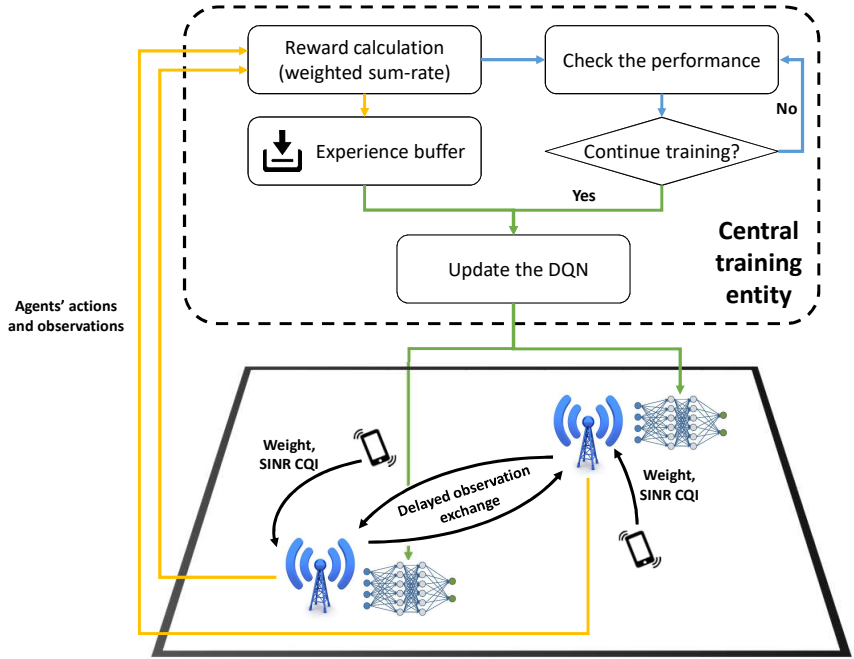}
\caption{Architecture diagram for training and inference in real-world deployments.}
\label{fig:real}
\end{figure}

\begin{itemize}
\item \textbf{Inference and training links (black):} These links represent the means by which the APs receive their observations in order to make their scheduling decisions. The APs rely on periodic weight and SINR CQI feedback from the UEs, as well as delayed observations from a set of neighboring APs. These links are used during inference as well as training.
\item \textbf{Training-only links (colored):} These links represent the information exchange between the APs and the central training entity, as well as the internal mechanism of the central training entity. These links are used only during training. They can further be partitioned into three subsets, depending on the rate of information exchange:
\begin{itemize}
\item \textit{Medium frequency links (orange)}: These are the links on which the observations and actions of the APs in the network are collected with a medium periodicity by the central training entity. Upon receiving the observations and actions, the central training entity can estimate a reward for each of the constituent time steps and then push the experiences to a central experience buffer, gathering the experiences of all APs.
\item \textit{Low frequency links (green)}: These links are used for updating the neural networks at the APs. For each training step, the central training entity fetches a random batch of samples from the experience buffer and uses them to calculate an updated set of weights for the DQN, which are then pushed to the agents at all APs. As the policy converges, these training updates can happen at a much slower time scale, e.g., only when there are significant changes in the environment and the model performance is lower than expected.
\item \textit{Ultra-low frequency links (blue)}: These links characterize training pause/termination check that happens on a much lower rate than the rest of the events. The central training entity needs to monitor the performance of the trained policy once in a while to determine whether or not to continue training. Training can be paused once the network performance satisfies certain criteria and otherwise, the training may continue.

\end{itemize}

\end{itemize}

The framework described above can be used to refine the policy of the agents to match the specific characteristics of the network in which they are deployed, and to adapt the policies to changes in these characteristics. But care must be taken to ensure that the policy is not skewed by statistically anomalous experiences, such as, for example, a large density of users in an area of the network that appear for a few days due to a large, temporary event. The issue of distributed fine-tuning of the agents during deployment is an interesting research topic and is left for future work.

\section{Concluding Remarks}
In this article, a framework for applying multi-agent deep reinforcement learning to the problem of link scheduling in downlink wireless networks is proposed in order to mitigate the interference between concurrent transmissions, while managing resources fairly across users experiencing different channel conditions. Each AP in the network is equipped with a deep Q-network (DQN) agent, making decisions whether the AP should be activated at each scheduling interval. Through a reward mechanism that encourages fair resource allocation, the agents interact with the environment by receiving local observations from the environment and remote observations from a subset of neighboring agents, and over time, they learn to improve the overall performance of the network. Simulation results demonstrate that the proposed approach outperforms baselines of full reuse and TDM in average rate and $5^{th}$ percentile rate, which is a measure of how fair the resource allocation is. Moreover, it is demonstrated that an agent trained on a network with four APs generalizes very well to networks with up to ten APs without sacrificing performance.

\pagebreak

\bibliographystyle{IEEEtran}
\bibliography{IEEEabrv,references}

% that's all folks
\end{document}